\documentclass[sn-apa,iicol]{sn-jnl}

\jyear{2021}%

\theoremstyle{thmstyleone}%
\theoremstyle{thmstyletwo}%

\theoremstyle{thmstylethree}%

\usepackage{hyperref}
\hypersetup{
	colorlinks=true,
	linkcolor = red,
	citecolor= blue,
	urlcolor=blue
}

\usepackage{amsmath,amssymb,amsfonts}%
\usepackage{array}
\usepackage[caption=false,font=normalsize,labelfont=sf,textfont=sf]{subfig}
\usepackage{textcomp}
\usepackage{stfloats}
\usepackage{url}
\usepackage{verbatim}
\usepackage{graphicx}
\usepackage{amsmath}
\usepackage{amssymb}
\usepackage{booktabs}
\usepackage{multirow}
\usepackage{amsthm}%
\usepackage{mathrsfs}%
\usepackage[title]{appendix}%
\usepackage{xcolor}%
\usepackage{textcomp}%
\usepackage{manyfoot}%
\usepackage{booktabs}%
\usepackage{algorithm}%
\usepackage{algorithmicx}%
\usepackage{algpseudocode}%
\usepackage{listings}%

\begin{document}

\title[Article Title]{Simplifying Low-Light Image Enhancement Networks with Relative Loss Functions}


\author[1]{\fnm{Yu} \sur{Zhang}}\email{zhangyuhit2@hit.edu.cn}
\author*[1]{\fnm{Xiaoguang} \sur{Di}}\email{dixiaoguang@hit.edu.cn}
\author[2]{\fnm{Junde} \sur{Wu}}\email{jundewu@ieee.org}
\author[1]{\fnm{Yue} \sur{Wang}}\email{seraph@hit.edu.cn}
\author[1]{\fnm{Yong} \sur{Li}}\email{522384058@qq.com}
\author[3]{\fnm{Rao} \sur{Fu}}\email{1098712828@qq.com}
\author[2]{\fnm{Yanwu} \sur{Xu}}\email{ywxu@ieee.org}
\author[1]{\fnm{Guohui} \sur{Yang}}\email{gh.yang@hit.edu.cn}
\author*[1]{\fnm{Chunhui} \sur{Wang}}\email{wang2352@hit.edu.cn}

\affil*[1]{\orgname{Harbin Institute of Technology}, \orgaddress{ \city{Harbin}, \postcode{150001}, \country{China}}}

\affil[2]{\orgname{South China University of Technology}, \orgaddress{\city{Guangzhou}, \postcode{510000},  \country{China}}}

\affil[3]{\orgname{Mind Vogue Lab}, \orgaddress{\city{Beijing}, \postcode{100000}, \country{China}}}


\abstract{Image enhancement is a common technique used to mitigate issues such as severe noise, low brightness, low contrast, and color deviation in low-light images. However, providing an optimal high-light image as a reference for low-light image enhancement tasks is impossible, which makes the learning process more difficult than other image processing tasks. As a result, although several low-light image enhancement methods have been proposed, most of them are either too complex or insufficient in addressing all the issues in low-light images. In this paper, to make the learning easier in low-light image enhancement, we introduce FLW-Net (Fast and LightWeight Network) and two relative loss functions. Specifically, we first recognize the challenges of the need for a large receptive field to obtain global contrast and the lack of an absolute reference, which limits the simplification of network structures in this task. Then, we propose an efficient global feature information extraction component and two loss functions based on relative information to overcome these challenges. Finally, we conducted comparative experiments to demonstrate the effectiveness of the proposed method, and the results confirm that the proposed method can significantly reduce the complexity of supervised low-light image enhancement networks while improving processing effect. The code is available at \url{https://github.com/hitzhangyu/FLW-Net}.}




\maketitle

\section{Introduction}\label{sec1}

Images captured in dark environments or with insufficient exposure often become low-light images that suffer from low contrast, low brightness, severe noise, and color deviation, making some information in the images invisible. To improve the quality of these images, numerous low-light image enhancement methods have been proposed in recent years \citep{guo2020zero, jiang2021enlightengan, zhang2021beyond, ma2022toward, liu2023TNNLS}. Although these methods have shown promising results, there remains a trade-off between processing speed and effect, regardless of whether they are based on learning or not.

\par While non-learning-based low-light image enhancement methods are capable of significantly improving image contrast and brightness\citep{guo2016lime, kimmel2003variational, Yu2019tcsvt, lee2013contrast}, subsequent separate denoising steps \citep{dabov2007image} or joint iterative denoising processes based on variation \citep{li2018structure} can be time-consuming. This limitation makes most of these methods unsuitable for real-time low-light image enhancement applications.  

\par Learning-based low-light image enhancement methods can be categorized into two main groups: the supervised learning-based and 
the unsupervised learning-based. Typically, unsupervised learning-based methods are lightweight and are more robust to various environments \citep{guo2020zero, li2021learning, ma2022toward}. However, they often lack responsive color correction and denoising methods, which limits their ability to improve the accuracy of subsequent high-level tasks \citep{better2021}. In contrast, supervised learning-based methods are designed to address all types of low-light image degradation and can significantly improve image quality \citep{Jiang2023tcsvt, Zhang2023tip, Zhou2023tim}. However, most of these methods require a complex network structure, resulting in longer processing times. 

\par Designing a lightweight network that can simultaneously enhance contrast and remove noise is a challenging task. Two primary challenges limit the simplification of the network. Firstly, the contrast adjustment of an image should consider both local and global information. This requires a large receptive field to capture global contrast, which increases the complexity of the network. Secondly, in low-light image enhancement, there is no absolute reference, which means the network must learn uncertain output values for the same input pixel or image block during denoising (We can name it the \textbf{one-to-many problem}). That makes it more difficult for supervised learning-based methods to learn compared to unsupervised learning-based methods, since most unsupervised learning-based methods have a consistent assumption about the output value, such as the average brightness value being close to a fixed value\citep{li2021learning}. 

\begin{figure}[htbp]
	\centering
	\includegraphics [width=\linewidth]{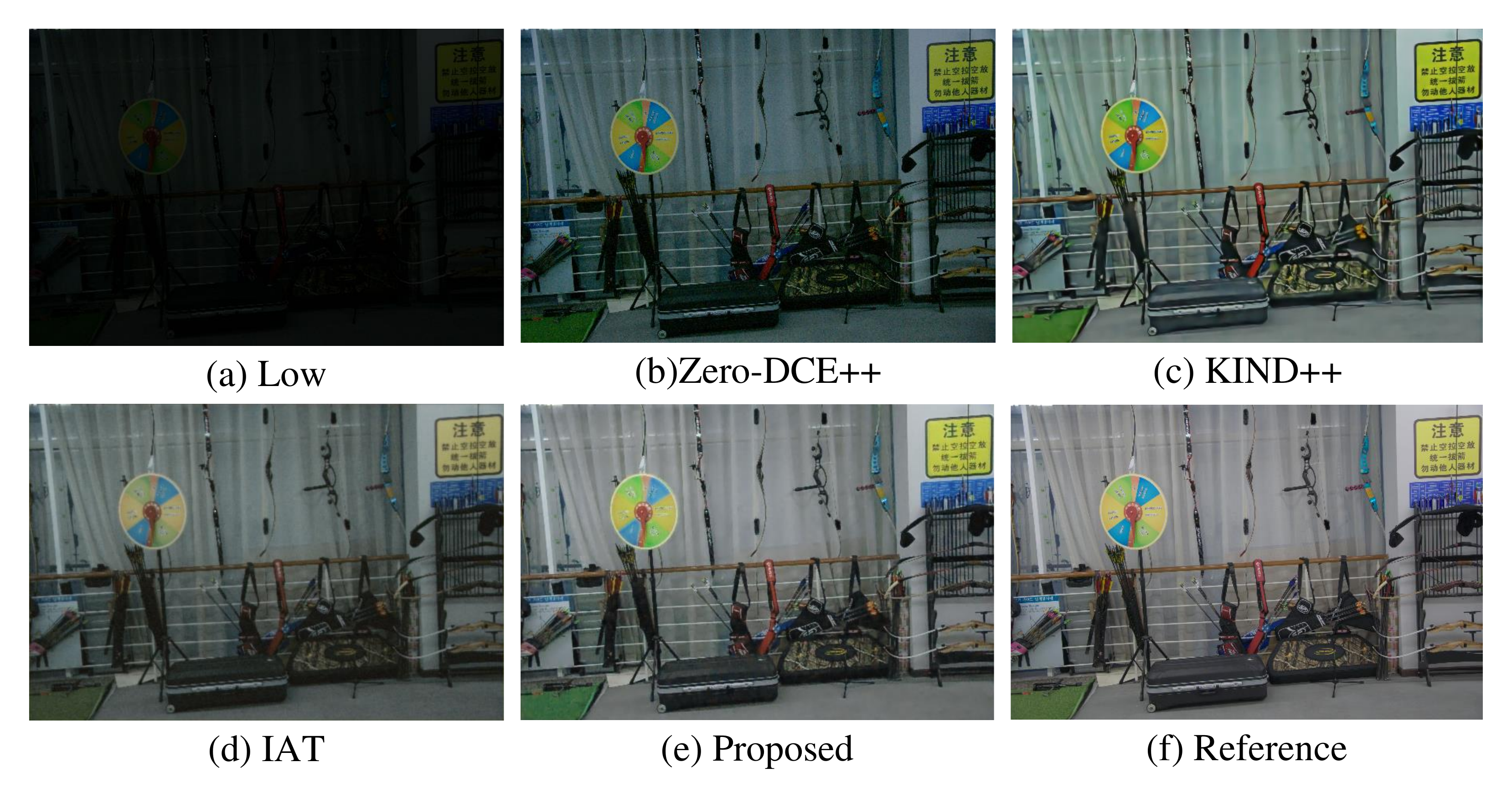}
	\caption{Visual comparison with some SOTA methods. (a) The original input low-light image. (b) to (e) are enhanced images produced by Zero-DCE++ \citep{li2021learning}, KIND++ \citep{zhang2021beyond} and the proposed method in this paper. (f) The reference image. It can be seen that the color of the enhanced image (e) produced by our proposed method is the closest to that of the reference image (f).}
	\label{fig_compare_1}
\end{figure}

\par In this paper, to address these challenges, we propose a solution by introducing a fast and lightweight image enhancement network called FLW-Net, along with two specially designed loss functions based on relative information for the low-light image enhancement task. We can abbreviate the losses based on relative information as the relative losses, and relative losses can achieve an extremum when the reference and the output are similar in some features, rather than requiring them to be exactly equal. An example of enhancing a low-light image comprising color deviation is shown in Fig. \ref{fig_compare_1}. It can be seen that the color of the enhanced image produced by proposed method is the closest to that of the reference image.

\par Specifically, in terms of network structure design, we propose a simple Global Feature Extraction (GFE) component that can extract global information from the image's histogram and generate a global brightness adjustment proposal through a higher-order curve adjustment method \citep{guo2020zero}. In terms of loss functions, we propose two losses based on relative information to alleviate the one-to-many problem in low-light image enhancement task.  

\par Our contributions can be summarized as follows:

\begin{itemize}
	
\item To improve computational efficiency by utilizing fewer parameters for obtaining global information, we proposed a Global Feature Extraction (GFE) component with only $1.4$K parameters. Unlike most existing methods that use large receptive fields or Transformer structures for extracting global information, GFE extracts information from histograms with a small number of bins. This not only makes it more efficient but also allows for easier integration with hyperparameters that control the degree of global enhancement.

\item To address the learning difficulty caused by multiple potential reference images, we proposed two novel loss functions ($L_{brightness}$ and $L_{structure}$) based on relative information. Unlike traditional $L_{1}$ or $L_{2}$ losses that aim to make the enhanced image exactly match the reference image, our proposed loss functions utilize cosine similarity, allowing for brightness differences between the output and reference. Then the network can focus more on color, structure and noise removal. When the proposed loss functions are combined with other supervised methods, it can achieve better quantitative metrics with less parameters(e.g., only about $5\%$ parameters of the original KIND’s network when combined with KIND \citep{zhang2019kindling}).

\item With GFE and two proposed loss functions, we built a fast and lightweight low-light image enhancement network, named FLW-Net, which achieves comparable or even better performance compared to state-of-the-art methods while maintaining faster processing speed. This shows the potential of reducing the complexity of low-light image enhancement networks with the proposed methods.

\end{itemize}
\section{Related Work}
\subsection{Low-Light Image Enhancement}
Non-learning-based low-light image enhancement solutions mainly include histogram equalization, gamma correction, methods based on dehazing \citep{dong2011fast} or the Retinex model \citep{wang2013naturalness}, as well as other improved methods based on these approaches \citep{celik2011contextual, guo2016lime, park2017low, fu2019hybrid, Kumar2022tcsvt}. Although these methods can significantly improve the brightness and contrast of images, removing noise and restoring color remain challenging.

\par Recently, learning-based methods for enhancing low-light images have achieved promising results, including supervised methods \citep{zhang2019kindling, zhang2021beyond, better2021, Xu2022tcsvt} and unsupervised methods \citep{guo2020zero, xiong2020unsupervised, jiang2021enlightengan, zhang2021self, ma2022toward, LIU2023109039pr}. However, unlike other image processing or computer vision tasks, the low-light image enhancement task usually lacks an  ground-truth/absolute label. For the same scene, there may be multiple low-light and high-light images, making it difficult to determine which reference image is the best. Even after expert correction, it can still be challenging to select the ideal reference image \citep{zhang2021beyond, better2021}.

\par Therefore, most supervised methods for low-light image enhancement often face challenges due to the presence of multiple potential reference images. To address this issue, there are mainly two types of methods. The first involves designing complex networks, such as the current state-of-the-art method MAXIM \citep{tu2022maxim}. While these methods can achieve promising visual results, they are often time-consuming for low-light image enhancement tasks.

\par The second type of method involves connecting the input and output during training by introducing hyperparameters \citep{chen2018learning, fu2020learning}, simplified Retinex models \citep{wei2018deep}, or both \citep{zhang2019kindling}. For instance, \cite{chen2018learning} introduced the exposure time ratio of reference and input images as a hyperparameter to achieve denoising and color restoration. \cite{fu2020learning} further proposed a sub-network to automatically select hyperparameters. \cite{wei2018deep} incorporated the simplified Retinex model into the network. However, the assumption in the simplified Retinex model that all three color channels have the same illumination image does not align with reality \citep{better2021}, leading to suboptimal denoising effects. 

\par To address this issue, \cite{zhang2019kindling} and \cite{zhang2021beyond} introduced both hyperparameters and the simplified Retinex model into the network and designed a separate restoration module to remove noise and correct color in the reflection image. However, it is important to note that the introduction of both Retinex models and hyperparameters might still deviate from the real imaging process. Consequently, while these methods involve less complex network structures, their running time may not meet the requirements of practical applications.

\par Unsupervised methods for low-light image enhancement typically assume that the output meets certain constraints, making them more lightweight and more stable for unseen scenes. For example, \cite{guo2020zero} use the mean value assumption (e.g., supposing the mean brightness of the image is between 0.4 and 0.6) and some specially designed loss functions to constrain the output. \cite{xiong2020unsupervised} specify the initial value of the illumination image in the simplified Retinex model as the max value of {R,G,B} in each pixel. \cite{jiang2021enlightengan} propose to learn constraints on the output from the normal-light images through GAN framework. \cite{ma2022toward} proposed to constrain the similarity of outputs at different stages during training. Although most of these unsupervised methods can meet real-time requirements, accurately removing noise and restoring color remains a challenge due to the lack of sufficient noise and color constraints.

\par By comparing supervised and unsupervised methods, it can be observed that if we do not aim to achieve enhancement results that are identical to the reference image but rather focus on learning color, structure, and noise removal from the reference image, it is possible to utilize a simple network to effectively remove noise and restore colors during image enhancement. At this point, the problem to be addressed is how to design evaluation metrics or loss functions that are not affected by brightness differences.

\subsection{Image Retouching}
Image retouching methods focus on problems such as inappropriate brightness, poor contrast, color deviation, etc., similar to image enhancement tasks \citep{wang2022neural}. However, most of these methods do not consider the noise problem. Therefore, basic retouching operations can work on a single pixel, making them extremely fast and lightweight \citep{he2020conditional, liu2022very, zeng2020learning, wang2022neural}. For example, \cite{he2020conditional} and \cite{liu2022very} proposed the Conditional Sequential Retouching Network (CSRNet) with only 37K trainable parameters. \cite{wang2022neural} proposed the trainable neural color operators, which contains only 28K parameters in their method. The successful applications of simple networks in image retouching also demonstrate the feasibility of using simple networks for image enhancement tasks.

\begin{figure*}[htpb]
	\centering
	\includegraphics[width=0.95\linewidth]{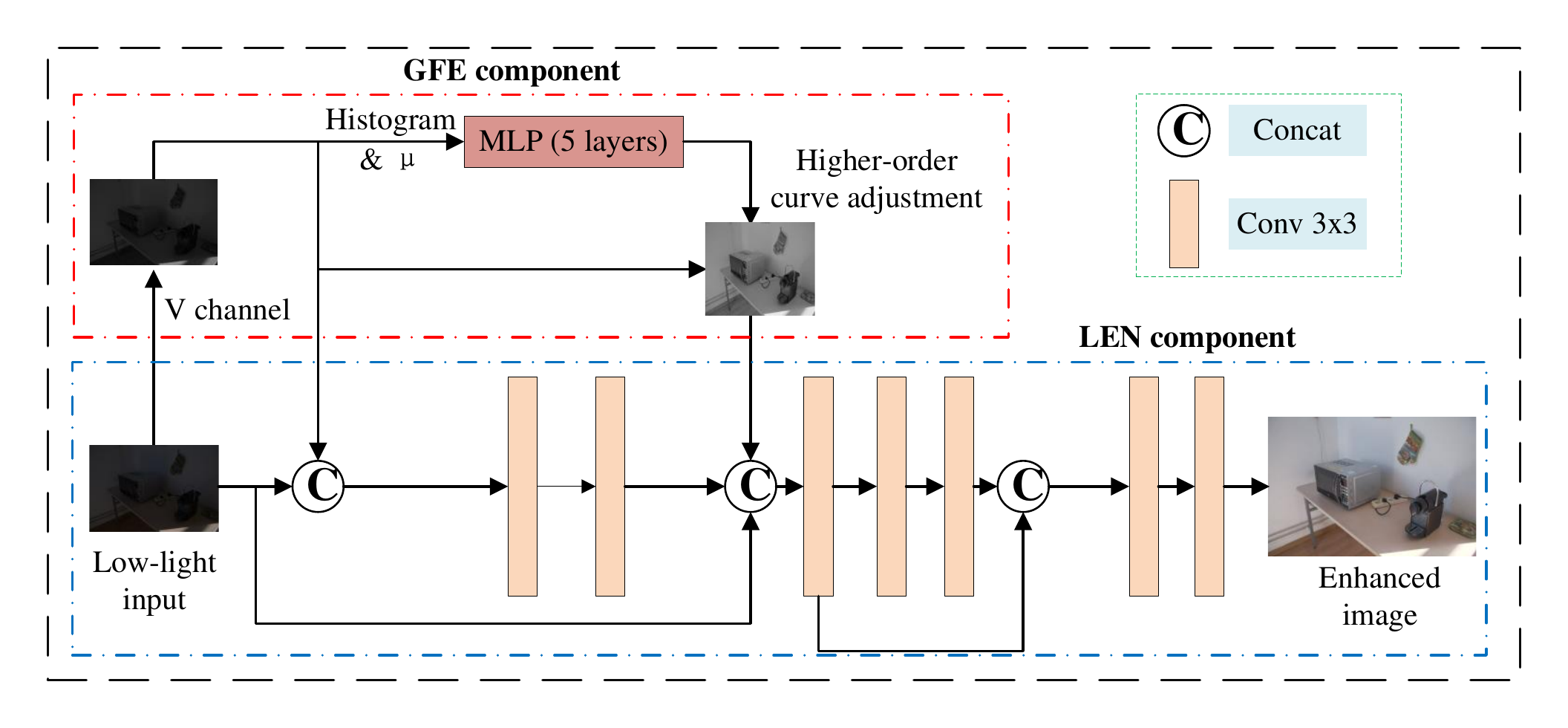}
	\caption{The detailed structure of the proposed method.}
	\label{figure_1_structure}
\end{figure*}

\section{Methodology}
Figure \ref{figure_1_structure} illustrates the architecture of FLW-Net, which comprises two primary modules: the Global Feature Extraction (GFE) component and the Local Enhancement Network (LEN) component. The GFE component takes the low-light image's V channel and the desired average brightness as inputs and produces a global brightness adjustment proposal through higher-order curve adjustment method. Then, the proposal is concatenated and fed into LEN. The LEN component takes the low-light image and the global brightness adjustment proposal as inputs and enhances the image with some carefully designed loss functions. It consists of several convolutional layers with a local receptive field to capture local information and generate high-frequency details.  

The proposed method includes several loss functions beyond the commonly used $L_{1}$ and SSIM. The color loss, denoted as $L_{color}$, is used to measure the color similarity between the enhanced image and the reference image. The brightness loss, denoted as $L_{brightness}$, is used to measure the difference in brightness orders between the enhanced image and the reference image. Finally, the structure loss, denoted as $L_{structure}$, is designed to encourage the enhanced image to have similar gradient orders to the reference image. We refer to these losses as relative losses. 

Unlike in $L_{1}$ and SSIM losses, where the extremum is only reached when the output and reference are exactly equal, the relative loss can achieve the extremum when the output and reference share some common features. All of these loss functions, including the $L_{1}$ and SSIM loss, are combined to form the total loss function used in training the FLW-Net. To the best of our knowledge, $L_{brightness}$ and $L_{structure}$ are first proposed by this paper in this low-light image enhancement task.

\subsection{Global Feature Extraction Component}
It is unnecessary to emphasize the importance of global information extraction in low-light image enhancement, as it has been extensively discussed in previous literature \citep{Cui2022BMVC}. However, the challenge lies in efficiently extracting global information and integrating it into the enhancement network.

\cite{better2021} has proven that the V channel in the HSV color space is sufficient to represent the brightness of the input low-light image. Meanwhile,  \cite{guo2020zero} proposed to iteratively apply Equation (\ref{eqn1}) to adjust the input low-light image. 

\begin{equation}
	\label{eqn1}
	\mathbf{I_{k+1}}=\mathbf{I_{k}}+\alpha_{k}\mathbf{I_{k}}(1-\mathbf{I_{k}}) 
\end{equation}
where $k$ represents the number of iterations (e.g., $\mathbf{I_{0}}$ represents the input low-light image).

\par Inspired by those two works, we propose to extract global information from the histogram of the V channel of the low-light image and represent it as higher-order curve coefficients, denoted as $\alpha_{0,1,...t}$. Specifically,
\begin{equation}
	\label{eqn2}
	\{\alpha_{0,1,...t}\}=G(H(\mathbf{I^{v}}))
\end{equation}
where $\mathbf{I^{v}}$ represents the V channel of the low-light image $\mathbf{I}$ in the HSV color space, $H(\mathbf{I^{v}})$ represents the operation of obtaining the histogram of the image $\mathbf{I^{v}}$, and $G(\cdot)$ represents the Global Feature Extraction (GFE) component, which is implemented using a Five-layer Multi-Layer Perception with only $1.4$K parameters. Compared with the method of \cite{guo2020zero}, the GFE component in our proposed method extracts global information from the histogram of the V channel instead of at the pixel level, which is more efficient. Specifically, the computational cost of the GFE component does not significantly increase with the increase in image size. Additionally, the output of GFE component is not the final enhanced result, which allows for further denoising and color restoration. Compared with the method of \cite{better2021}, the proposed method does not need to be combined with other image contrast enhancement methods and can achieve end-to-end training with the help of the loss functions proposed in Eqn. (\ref{eqn9}) and Eqn. (\ref{eqn11}). 

As mentioned, there is currently no definitive standard that defines whether an enhancement result is optimal. Similarly, we cannot claim that the processing result of a fixed-parameter network trained on a specific dataset is optimal. In this case, introducing adjustable parameters for image enhancement is a reasonable solution, as adopted in previous work \citep{liu2023TNNLS, zhang2021beyond, wang2022llflow}. Additionally, if the parameters are related to reference images, we can allow the network to learn relatively deterministic mapping operations, thereby simplifying the training difficulty of the network in the training phase.

The GFE component can be easily modified to incorporate additional hyperparameters. In this paper, we used the desired average brightness value, $\mu$, as the hyperparameter. During the training process, $\mu$ is calculated as the mean value of the V channel of the reference image. Compared to previous works which utilized parameters such as exposure time difference between reference images and input \citep{chen2018learning}, and brightness ratio between the illumination image of the reference image and that of the input image \citep{zhang2019kindling}, $\mu$ is more intuitive and highly correlated with histogram information. Moreover, during the testing process, we can easily fix $\mu$ to a constant value, as done in some unsupervised works\citep{guo2020zero}. Therefore, the modified GFE component can be expressed as:
\begin{equation}
	\label{eqn3}
	\{\alpha_{0,1,...t}\}=G(H(\mathbf{I^{v}}),\mu)
\end{equation}
\par Then, we can iteratively adjust the V channel of the low-light image to obtain the global brightness adjustment proposal, as follows:
\begin{equation}
	\label{eqn4}
	\mathbf{I^{v}_{k+1}}=\mathbf{I^{v}_{k}}+\alpha_{k}\mathbf{I^{v}_{k}}(1-\mathbf{I^{v}_{k}}) 
\end{equation}

After obtaining the global brightness adjustment proposal, it will be concatenated with the middle layer of LEN and fed into the LEN for further enhancement. The entire FLW-Net is trained end-to-end, which means that all the components are trained jointly to optimize the overall performance of the network.

\subsection{Loss Functions based on Relative Information}
Let us consider $\{\mathbf{I},\mathbf{Y}\}$ as one paired low/high-light images. Typically, we use lots of paired images to train the enhancement network $E$ and hope it can well fit the following Equation (\ref{eqn5}):
\begin{equation}
	\label{eqn5}
	\mathbf{Y_{(i,j)}} = E(\mathbf{I_{(i,j)}})
\end{equation}
where $(i,j)$ represents the coordinate of one pixel. To achieve this, various loss functions have been adopted to train the enhancement network $E$ with paired low/high-light images, including commonly used $L_1$, $L_2$ and SSIM loss, etc. However, due to the one-to-many problem, these loss functions may not be as effective for image enhancement as they are for other low-level image processing tasks such as image denoising, image deblurring, and image dehazing. In other words, these loss functions are more suitable for learning the mapping relationship with an absolute reference image for the input.

\par As previously mentioned, previous studies have proposed strategies to establish a one-to-one relationship between the input and output of the enhancement models (where one input low-light image corresponds to one certain reference image). However, the Retinex models and additional parameters adopted in those work cannot onvert the one-to-many problem into a real one-to-one problem. In practice, even adding a hyperparameter for each pixel does not guarantee a satisfactory color restoration effect\citep{better2021}. Therefore, most of previous works often require complex networks or separate denoising modules, such as SID \citep{chen2018learning}, RetinexNet \citep{wei2018deep}, KinD \citep{zhang2019kindling}, KinD++ \citep{zhang2021beyond}. 

\par Since there is no absolute supervision information, one intuitive approach is to use relative information in the loss functions, which reduces the assumption of existence of absolute reference images. Previous unsupervised methods have proposed some useful loss functions, such as spatial consistency loss \citep{guo2020zero}, normalized gradient error \citep{zhang2021self}, and perception loss \citep{jiang2021enlightengan}. However, most of them do not achieve impressive results in supervised methods, since they impose weak constraints.

\par In this paper, we propose the use of relative losses, which are loss functions based on relative information, to make the learning easier for networks.

Firstly, for one pixel, we expect its color information more than the brightness to match the reference image. To achieve this, we need to extract the image's color information, which can be accomplished through various color spaces, such as the HSV (Hue, Saturation, Value) or HSI (Hue, Saturation, Intensity) color space. In this paper, we chose to use the HSV color space. It has been demonstrated that two pixels share the same Hue and Saturation when they satisfy the following equation \citep{better2021}:
\begin{equation}
	\label{eqn6}
	E\mathbf{(I_{(i,j)})} = \lambda \mathbf{Y_{(i,j)}}
\end{equation}
where $\lambda$ represents arbitrary non-zero positive number. It should be noticed that both $E\mathbf{(I_{(i,j)})}$ and $ \mathbf{Y_{(i,j)}}$ represent 3D vectors. Then, we can adopt the cosine similarity to measure the Hue and Saturation difference between two pixels. Therefore, the loss function for color can be designed as follows:
\begin{equation}
	\label{eqn7}
	L_{color} = 1 - \sum_{i=1,j=1}^{m,n} <E\mathbf{(I_{(i,j)})}, \mathbf{Y_{(i,j)}}>
\end{equation}
\noindent where $<\cdot,\cdot>$ represents the cosine similarity of two vectors. By minimizing this loss function, the network is encouraged to match the Hue and Saturation information between the output and reference images. This loss function has also been adopted in many other image enhancement \citep{Liu2022Tcsvt,Zhang2023tip} and Computational Color Constancy work \citep{1036047,5719167}.

Secondly, regarding the brightness, it is expected that the enhanced images have the same lightness order as the reference \citep{wang2013naturalness}, which means that images that are brighter in the reference should also be brighter in the enhanced image. \cite{wang2013naturalness} proposed the evaluation metric, LOE (Lightness-Order-Error), for the lightness order error. However, directly incorporating LOE into the loss function is not straightforward since it is difficult to calculate the corresponding gradients in back propagation. In this paper, we propose the following equation to model the brightness relation between the enhanced image and the reference.
\begin{equation}
	\label{eqn8}
	b(E \mathbf{(I_{(i,j)})}) = \beta b(\mathbf{Y_{(i,j)}}) + \gamma 
\end{equation}
\noindent where $b(\cdot)$ represent image blocks centered on pixels $E \mathbf{(I_{(i,j)})}$ and $\mathbf{Y_{(i,j)}}$, $\beta$ and $\gamma$ can represent 3D vectors for color image or scalars in gray images. For different blocks, there can be different $\beta$ and $\gamma$. This is useful for processing images with non-uniform brightness, where different regions of the image have different enhancement levels. It can be seen that, in this model, the enhanced images have the same lightness order as the reference image in every image block. Also, it is more rigorous than the LOE metric since it requires more than just brightness order. Specifically, it requires the enhanced image to be a linear transformation of the noise-free reference image, which significantly suppresses noise. Then, we can design the loss function as follows: 
\begin{align}
	L_{brightness} &= 1 - \sum_{c \in{{R,G,B}}}^{} \sum_{i=1,j=1}^{m,n} \nonumber\\&<b(E \mathbf{(I^{c}_{(i,j)})})- min(b(E \mathbf{(I^{c}_{(i,j)})})), \nonumber\\&b(\mathbf{Y^{c}_{(i,j)}})-min(b(\mathbf{Y^{c}_{(i,j)}}))> 
	\label{eqn9}
\end{align}
\noindent where $c$ represents the different color channels, and the purpose of subtracting the minimum value is to remove the influence of the constant $\gamma$. The following experimental results show that $L_{brightness}$ can improve the PSNR metric which is related to noise suppression(In Sec. \ref{Ablation_study}, Table \ref{table_psnr_ssim} and Figure \ref{fig_psnr_ssim}).

Thirdly, for the structure information, it is usually expressed by gradient information. We can adopt a similar model as Equation (\ref{eqn8}). The difference is replacing the brightness value with gradient. Then, we can get the following Equation (\ref{eqn10}).
\begin{equation}
	\label{eqn10}
	b(\nabla E \mathbf{(I_{(i,j)})}) = \eta b(\nabla \mathbf{Y_{(i,j)}}) + \epsilon 
\end{equation}

Then, the loss funciton for structure can be expressed as follows:
\begin{align}
	L_{structure} &= 1 - \sum_{c \in{{R,G,B}}}^{} \sum_{i=1,j=1}^{m,n} \nonumber\\&<b(\nabla E \mathbf{(I^{c}_{(i,j)})})- min(b(\nabla E \mathbf{(I^{c}_{(i,j)})})), \nonumber\\&b(\nabla \mathbf{Y^{c}_{(i,j)}})-min(b( \nabla \mathbf{Y^{c}_{(i,j)}}))> 
	\label{eqn11}
\end{align}

\par If we remove the $\epsilon$ part in Eqn.(\ref{eqn10}), Eqn.(\ref{eqn10}) will become the derivative of the Eqn.(\ref{eqn8}). At this point, $L_{brightness}$ and $L_{structure}$ will essentially represent the same thing. However, by introducing $\epsilon$ in Eqn.(\ref{eqn10}), it will allow for non-linear changes in brightness within image blocks. This more closely aligns with the imaging process in reality and can reduce the effects of saturation and under-saturation caused by linear changes in brightness during enhancement. As a result, $L_{structure}$ is more effective in preserving image structures. Experimental results also indicate that it is more effective in improving the structural similarity index (SSIM) metric(In Sec. \ref{Ablation_study}, Table \ref{table_psnr_ssim} and Figure \ref{fig_psnr_ssim}).

The total loss can be expressed as follows:
\begin{equation}
	\label{eqn12}
	L_{ALL} = L_{1} + L_{SSIM} + L_{color} + L_{brightness} + L_{structure}
\end{equation}
where $L_{SSIM}$ represents the SSIM loss between the enhanced and reference images.

\begin{figure*}[htpb]
	\centering
	\includegraphics[width=0.93\linewidth]{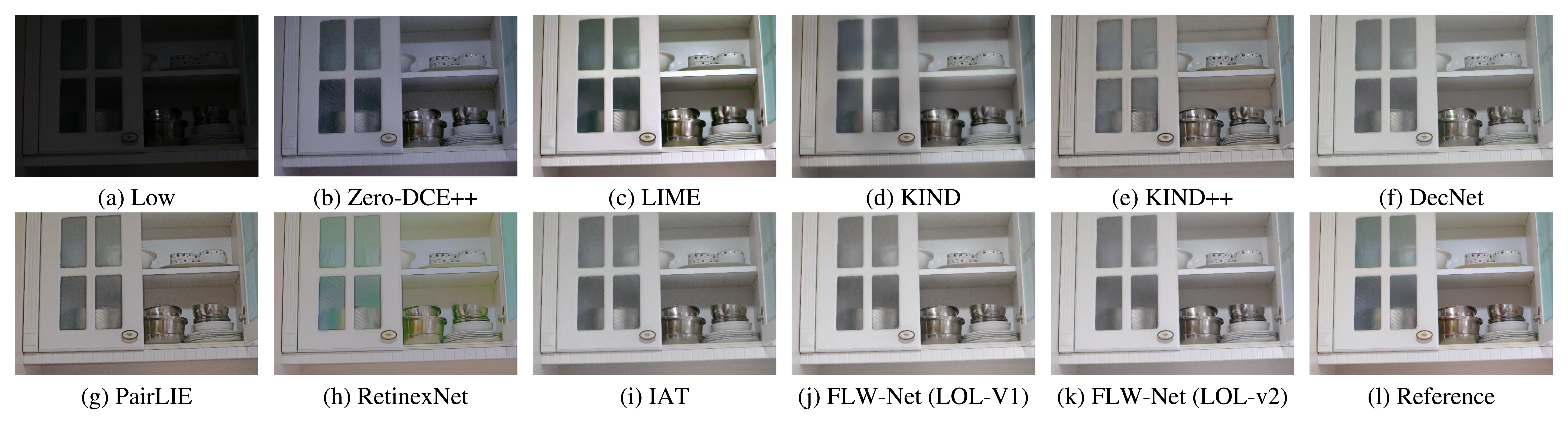}
	\caption{Visual comparison results on LOL-V1 dataset. Comparison methods include Zero-DCE++, LIME, KIND, KIND++, DecNet, PairLIE, RetienxNet, IAT and the proposed FLW-Net trained on LOL-V1 and LOL-V2 datasets. (Best viewed on high-resolution displays with zoom-in.)
	}
	\label{figure_comare_2}
\end{figure*}

\section{Experiments}
\subsection{Implementation Details}
The framework is implemented with PyTorch on an NVIDIA 3090 Ti GPU. The batch size used for training is 171. We use the Adam optimizer to train the network with a learning rate of 0.0001. We mainly use two datasets for training and testing: the LOL-V1 dataset \citep{wei2018deep} and the LOL-V2 dataset \citep{yang2021sparse}. The LOL-V1 dataset contains 500 image pairs, with 15 pairs used for testing. The LOL-V2 dataset contains 689 image pairs for training and 100 pairs for testing. We also collected some images online and merged them together, which we refer to as the Mixed dataset. The Mixed dataset includes the test images of LOL-V1 (15 images), LIME \citep{guo2016lime} (10 images), MF \citep{fu2016fusion} (10 images), and VV\footnote{https://sites.google.com/site/vonikakis/datasets/challenging-dataset-for-enhancement} (23 images), and most of these images do not have corresponding reference images. 

We used four metrics for quantitative comparison: PSNR, SSIM, CIEDE2000 \citep{luo2001development,sharma2005ciede2000}, and NIQE \citep{mittal2012making}. PSNR and SSIM are reference image quality assessment methods that indicate the noise level and the structural similarity between the enhanced images and the reference, respectively. CIEDE2000 is a reference image quality assessment method to accurately measure colors differences, which is published by International Commission on Illumination(Also known as the Commission Internationale de l´Eclairage, CIE), and a lower value indicates less color difference. NIQE is a non-reference image quality assessment method that evaluates the naturalness of the image, and a lower value indicates better quality. To differentiate between the $\mu$ values during the training and testing processes, we use $\mu_{train}$ and $\mu_{test}$ to represent them, respectively.

\subsection{Objective Performance Evaluation}
\begin{figure*}[htpb]
	\centering
	\includegraphics[width=0.93\linewidth]{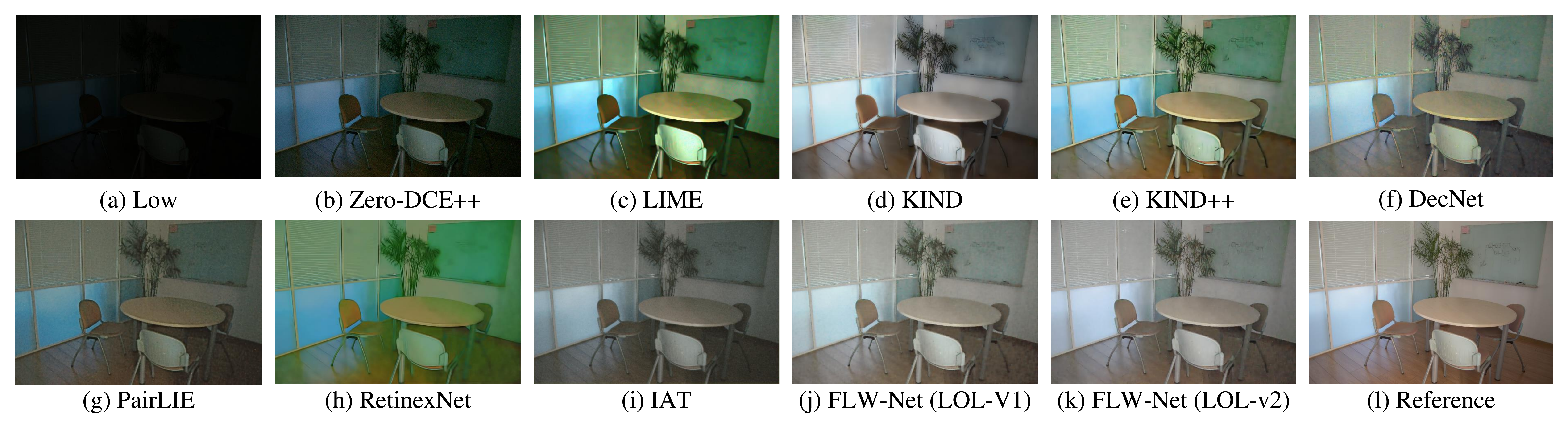}
	\caption{Visual comparison results on LOL-V2 dataset. Comparison methods include Zero-DCE++, LIME, KIND, KIND++, DecNet, PairLIE, RetienxNet, IAT and the proposed FLW-Net trained on LOL-V1 and LOL-V2 datasets. (Best viewed on high-resolution displays with zoom-in.)
	}
	\label{figure_comare_3}
\end{figure*}

\begin{figure*}[htpb]
	\centering
	\includegraphics[width=0.93\linewidth]{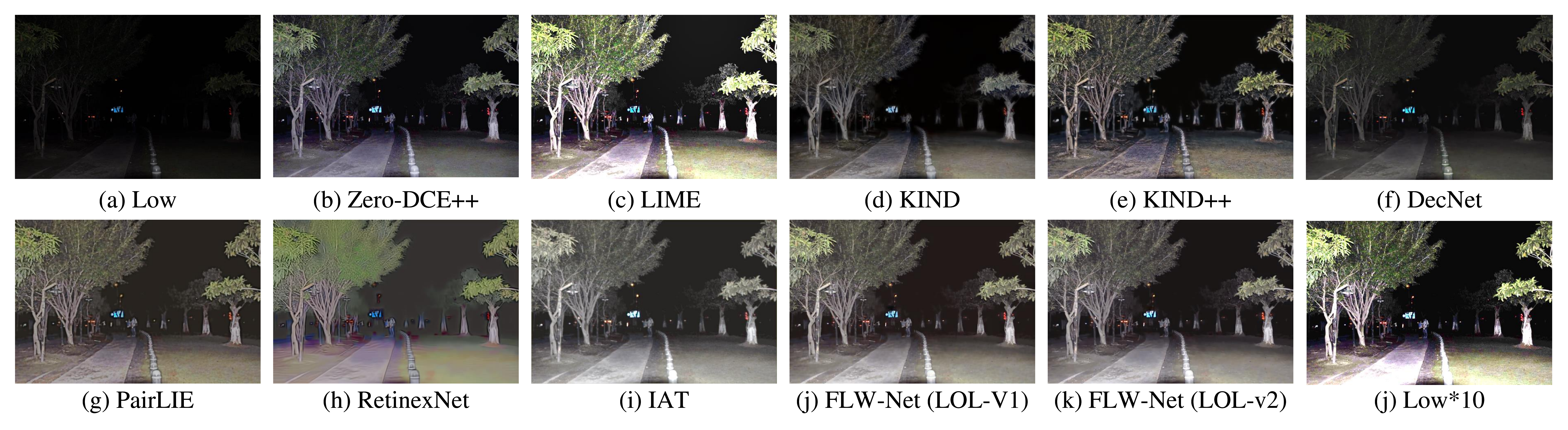}
	\caption{Visual comparison results on MF dataset \citep{fu2016fusion}. Comparison methods include Zero-DCE++, LIME, KIND, KIND++, DecNet, PairLIE, RetienxNet, IAT and the proposed FLW-Net trained on LOL-V1 and LOL-V2 datasets. (Best viewed on high-resolution displays with zoom-in.)
	}
	\label{figure_comare_4}
\end{figure*}

\begin{figure*}[htpb]
	\centering
	\includegraphics[width=0.93\linewidth]{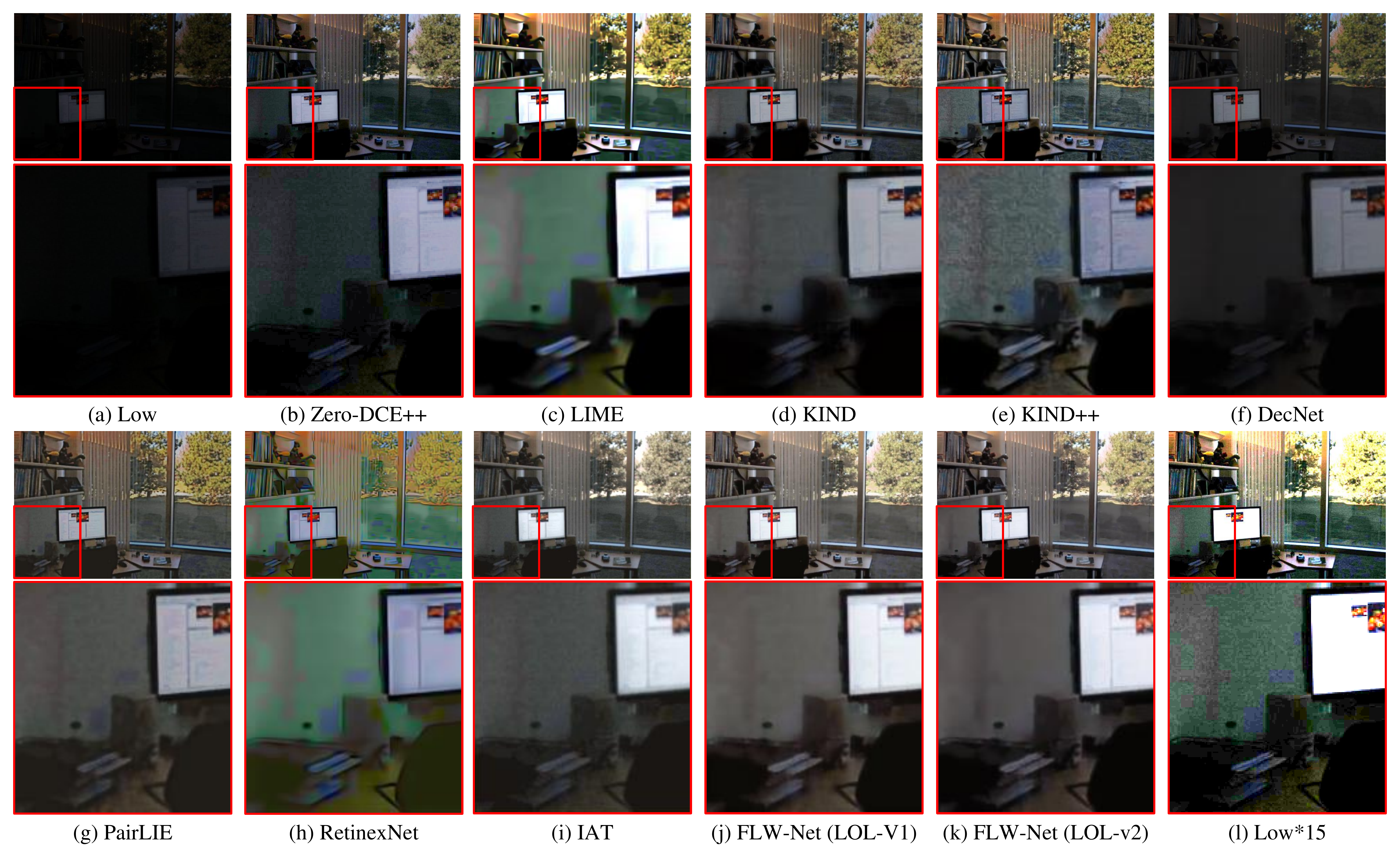}
	\caption{Visual comparison results on image from Internet. Comparison methods include Zero-DCE++, LIME, KIND, KIND++, DecNet, PairLIE, RetienxNet, IAT and the proposed FLW-Net trained on LOL-V1 and LOL-V2 datasets. (Best viewed on high-resolution displays with zoom-in.)
	}
	\label{figure_comare_7}
\end{figure*}

In this section, we compared our method with several state-of-the-art (SOTA) low-light image enhancement methods, including LIME \citep{guo2016lime}, RetinexNet \citep{wei2018deep}, Zerodce++ \citep{li2021learning}, KIND \citep{zhang2019kindling}, KIND++ \citep{zhang2021beyond}, DecNet \citep{liu2023TNNLS}, PairLIE \citep{fu2023learning} and IAT \citep{Cui2022BMVC}. Among them, LIME is a non-learning-based method, and ZeroDCE++ can be trained without any references. PairLIE can be trained with paired low-ligh images. The other methods are based on supervised learning. Among them, KIND \citep{zhang2019kindling}, KIND++\citep{zhang2021beyond} and DecNet\citep{liu2023TNNLS} all have hyperparameters during training and testing. The comparison results are shown in Table \ref{table_sota}, \ref{table_color_sota} and \ref{table_fixed_sota}  and Figures \ref{figure_comare_2}, \ref{figure_comare_3}, \ref{figure_comare_4} and \ref{figure_comare_7}. In Fig. \ref{figure_comare_4} and \ref{figure_comare_7}, the hyperparameter $\mu$ is fixed at a constant value of 0.4.

During the training on the LOL-V1 dataset, we utilized only 343 images, which is approximately half of the LOL-V2 training data. It should be noted that both datasets were produced by the same team, and LOL-V2 contains most of the data in LOL-V1. Hence, we can evaluate the impact of training data volume on the network.

As shown in Table \ref{table_sota}, the training data volume has a greater impact on PSNR than SSIM. For instance, when trained on the LOL-V1 dataset, the PSNR reduced by almost 0.9 dB compared to the LOL-V2 dataset (PSNR: $26.61 \rightarrow 25.71$). However, the SSIM only decreased by 0.01 (SSIM: $0.88 \rightarrow 0.87$). Furthermore, irrespective of whether trained on the LOL-V1 or LOL-V2 dataset, FLW-Net outperforms other methods in terms of PSNR. Regarding SSIM, FLW-Net, KIND, and KIND++ achieve similar results. However, FLW-Net has fewer parameters (only ~17K parameters) and consumes less running time during testing. It should be noted that when testing on the LOL-V2 and LOL-V1 datasets, the hyperparameters of DecNet, KIND and KIND++ are derived from the reference image.

\begin{table*}[htpb]
	\centering
	\caption{Quantitative comparison results on LOL (V1 \citep{wei2018deep} \& V2 \citep{yang2021sparse}) datasets and a mixed dataset. The mixed dataset includes the test images of LOL-V1 (15 images), LIME \citep{guo2016lime} (10 images), MF \citep{fu2016fusion} (10 images), and VV (23 images), and most of these images do not have corresponding reference images. It should be noted that the test time of KIND and KIND++ comes from KIND++\citep{zhang2021beyond} with a Titan-X GPU, and the test time of RetinexNet does not include the running time of BM3D\citep{dabov2007image} for denoising.}
	\label{table1}
	\resizebox{\textwidth}{!}{
		\begin{tabular}{c|ccc|ccc|c|cc}
			\hline 
			\multirow{2}{*}{Method}  & \multicolumn{3}{c|}{LOL-V1} & \multicolumn{3}{c|}{LOL-V2}& Mixed Dataset(Unpaired) & \multicolumn{2}{c}{Efficiency}\\
			\cline{2-10}
			\multicolumn{1}{c|}{} &PSNR$\uparrow$ &SSIM$\uparrow$ &NIQE$\downarrow$ &PSNR$\uparrow$ &SSIM$\uparrow$ &NIQE$\downarrow$  &NIQE$\downarrow$ & Params(M)$\downarrow$ & test time(s)$\downarrow$  \\
			\hline 
			LIME\citep{guo2016lime}        &17.22  &0.50 &5.32  &15.77  &0.46 &5.37 &4.57 &-  &0.190   \\  
			RetinexNet\citep{wei2018deep}    &17.86  &0.78 &6.37  &17.37  &0.76 &9.09  &5.68 &0.4  &0.019$^{*}$   \\  
			Zerodce++\citep{li2021learning} &15.35  &0.57 &7.86  &18.49  &0.58 &8.05 &4.53  &\textbf{0.01}  &\textbf{0.001}   \\
			KIND\citep{zhang2019kindling}   &20.38  &0.83 &5.45  &23.78  &\textbf{0.88} &4.96 &3.87 &8.21  & 0.11$^{*}$   \\ 
			KIND++\citep{zhang2021beyond}   &21.80  &\textbf{0.84} &5.17  &22.21  &0.84 &4.89 &3.74 &8.28  & 0.12$^{*}$   \\ 
			IAT\citep{Cui2022BMVC}       &23.38  &0.81 &\textbf{3.92}  &23.50  &0.82 &4.29 &4.71 &0.09  &0.004    \\ 
			DecNet\citep{liu2023TNNLS}	&22.49 &0.82 &4.51 &22.56 &0.84 &4.83 &4.26 &1.83 &0.353	\\
			PairLIE\citep{fu2023learning}	&18.47 &0.75 &4.25 &19.88 &0.78 &4.34 &3.90 &0.33 &0.057	\\
			FLW(Training on LOL V1)     &23.84  &0.83 &4.22  &25.71  &0.87 &4.09  &3.93 &0.02  &0.001    \\ 
			FLW(Training on LOL V2)     &\textbf{24.70}  &\textbf{0.84} &4.11  &\textbf{26.61}  &\textbf{0.88} &\textbf{3.89}  &\textbf{3.72}  &0.02  &\textbf{0.001}    \\ 
			\hline 
		\end{tabular}
	}
	\label{table_sota}
\end{table*}

\begin{table}[htpb]
	\centering
	\caption{Color comparison results on LOL (V1 \citep{wei2018deep} \& V2 \citep{yang2021sparse}) datasets. }
	\label{table1}
	\resizebox{0.45\textwidth}{!}{
		\begin{tabular}{c|cc}
			\hline 
			\multirow{2}{*}{Method} & \multicolumn{2}{c}{CIEDE2000$\downarrow$} \\
			\cline{2-3}
			\multicolumn{1}{c|}{}  & \multicolumn{1}{c}{LOL-V1} & \multicolumn{1}{c}{LOL-V2}   \\
			\hline 
			LIME\citep{guo2016lime}         &14.62 &17.63 \\  
			RetinexNet\citep{wei2018deep}   &13.76 &18.07\\  
			Zerodce++\citep{li2021learning} &19.28 &14.27  \\
			KIND\citep{zhang2019kindling}   &9.68  &6.74\\ 
			KIND++\citep{zhang2021beyond}   &8.51  &9.36\\ 
			IAT\citep{Cui2022BMVC}       &7.97  &8.17\\ 
			DecNet\citep{liu2023TNNLS}   &8.93   &8.87 \\
			PairLIE\citep{fu2023learning}  &11.93   &11.23 \\
			FLW(Training on LOL V1)    &7.74 &6.64\\ 
			FLW(Training on LOL V2)    &\textbf{7.23} &\textbf{6.06} \\ 
			\hline 
		\end{tabular}
	}
	\label{table_color_sota}
\end{table}

Table \ref{table_color_sota} presents the color comparison results of various methods on the LOL (V1 \citep{wei2018deep} $\&$ V2 \citep{yang2021sparse}) datasets. We used the total color difference $\Delta E_{00}$ in CIEDE2000 to evaluate different methods. As can be observed, the proposed method achieved lower CIEDE2000($\Delta E_{00}$) values than other methods, demonstrating the effectiveness of our proposed method in color restoration.

\begin{figure*}[htpb]
	\centering
	\includegraphics[width=0.93\linewidth]{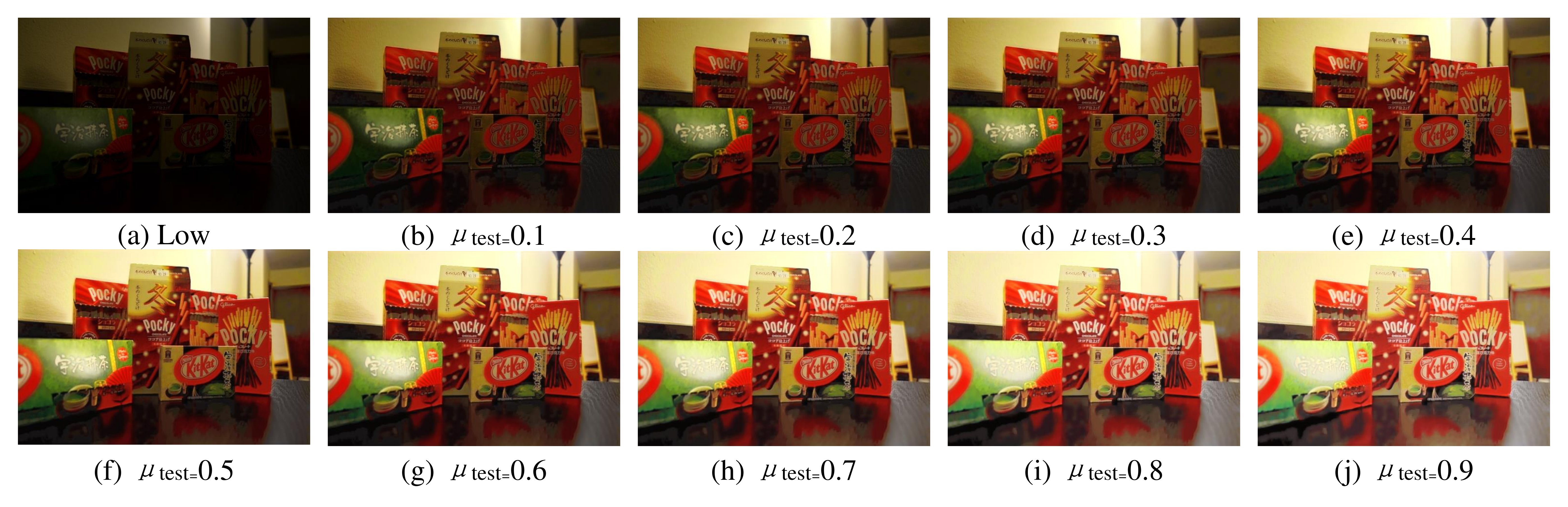}
	\caption{Visual comparison results on MF dataset when we change the $\mu_{test}$ value from $0.1$ to $0.9$ with a step of 0.1. (Best viewed on high-resolution displays with zoom-in.)}
	\label{figure_comare_5}
\end{figure*}

\begin{figure*}[htpb]
	\centering
	\includegraphics[width=0.93\linewidth]{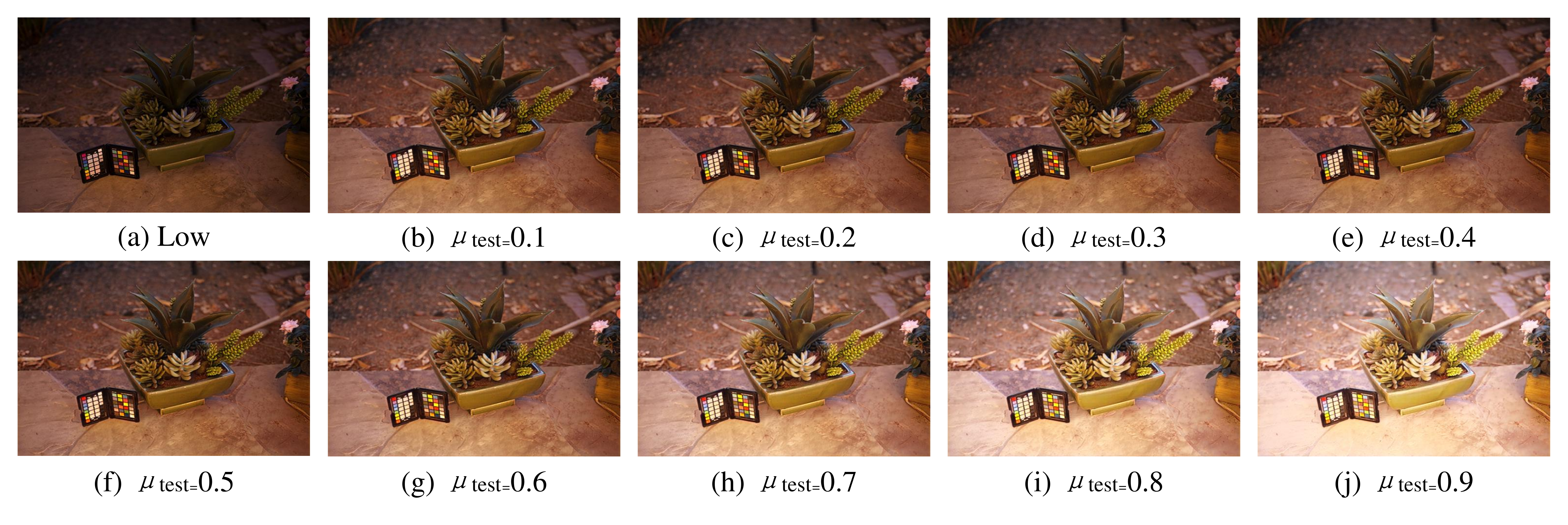}
	\caption{Visual comparison results on LIME dataset when we change the $\mu_{test}$ value from $0.1$ to $0.9$ with a step of 0.1.(Best viewed on high-resolution displays with zoom-in.)
	}
	\label{figure_comare_6}
\end{figure*}

\begin{table}[htpb]
	\centering
	\caption{Comparison results on LOL-V2 \citep{yang2021sparse}) datasets with fixed hyperparameter in all methods during test. ($\mu_{test} = 0.4$ is adopted in our method)}
	\label{table1}
	\resizebox{0.5\textwidth}{!}{
		\begin{tabular}{c|cccc}
			\hline 
			\multirow{1}{*}{Method$\uparrow$}  & \multicolumn{1}{c}{PSNR$\uparrow$} & \multicolumn{1}{c}{SSIM} & \multicolumn{1}{c}{NIQE$\downarrow$} & \multicolumn{1}{c}{CIEDE2000$\downarrow$}    \\
			\hline 
			KIND\citep{zhang2019kindling}   &20.59  &0.82 &4.86 &9.3872\\ 
			KIND++\citep{zhang2021beyond}   &17.66  &0.77 &4.73 &13.89\\ 
			DecNet\citep{liu2023TNNLS}     &21.13   &0.83 &4.85 &10.45 \\
			FLW(Ours)    &\textbf{23.50} &\textbf{0.86} &\textbf{3.88} &\textbf{8.28} \\ 
			\hline 
		\end{tabular}
	}
	\label{table_fixed_sota}
\end{table}

In Figures \ref{figure_comare_2} and \ref{figure_comare_3}, we can observe that the images enhanced by FLW-Net are more closely aligned with the reference images in terms of brightness, contrast, and color. The models used in LIME and RetinexNet are simplified Retinex models. Therefore, the Saturation and Hue of the enhanced images are identical to those of the original low-light images, especially in Figure \ref{figure_comare_3}. Although KIND and KIND++ introduced a restoration network to recover the color and remove noise in the reflection image, the results are still not stable. For instance, in Figure \ref{figure_comare_3}(e), the image enhanced by KIND++ still has color deviations compared with the reference image. In Figure \ref{figure_comare_4}(d), the small light source was treated as noise and removed by KIND. 

During testing, the value of $\mu_{test}$ can be obtained from reference images such as \citep{zhang2019kindling}, \citep{liu2023TNNLS}, and \citep{zhang2021beyond}. However, in practical applications, the value of $\mu_{test}$ is typically determined by the user or machine to dynamically adjust enhancement results, rather than relying on the reference image. Therefore, it is important to carefully consider the impact of $\mu_{test}$ values. In this regard, we also present the results obtained when $\mu_{test}$ takes on constant values during testing.

Table \ref{table_fixed_sota} shows the quantitative comparison results with other methods which also introduces hyperparameter, and it can be seen that the proposed method can achieve better performance than other methods with less parameters and simpler network structure. Fig. \ref{figure_comare_4} and \ref{figure_comare_7} shows the visual comparison results with other methods. It can be seen that, the proposed method can effectively remove noise while preserving structural information. 



\subsection{Ablation Study}
\label{Ablation_study}
We performed two ablation studies on the LOL-V2 dataset to demonstrate the effectiveness of each component in our proposed method. For quantitative comparison, we used the evaluation metrics of PSNR, SSIM and CIEDE2000.

\noindent\textbf{Contribution of Each Loss:} In this ablation study, we used the complete network trained with $L_{1}$ and SSIM loss as the baseline model. We then incorporated the relative loss functions into the network's loss function and retrained it to examine their effects on the performance of the model. Additionally, we also trained the network solely with the relative loss functions to evaluate their effectiveness. 


\par In Table \ref{table_psnr_ssim}, it can be observed that the addition of each relative loss to the baseline model leads to a slight improvement in PSNR or SSIM when $\mu_{test}$ is obtained from reference images. However, when $\mu_{test}$ takes on a constant value for all testing images, both $L_{brightness}$ and $L_{structure}$ demonstrate significant improvements in PSNR and SSIM. This demonstrates that the proposed two loss functions can help the network better learn noise removal and structural preservation, enhancing the stability of the network. Between these two losses, $L_{brightness}$ shows a better improvement in PSNR, which is related to its denoising ability, while $L_{structure}$ shows a better improvement in SSIM, which is related to its ability to retain structural information. On the other hand, the improvement in PSNR and SSIM with $L_{color}$ is relatively minor. This could be because $L_{color}$ only considers the information of a single pixel, whereas noise removal and retention of structural information require the introduction of information from surrounding pixels. 

\begin{table*}
	\centering
	\caption{The influence of different training losses. During testing, the input $\mu_{test}$ value can be a constant (e.g., $\mu_{test} = 0.4$) or obtained through the reference(e.g., $\mu_{test}$ equals the mean value of the reference's V channel).}
	\resizebox{0.8\textwidth}{!}{
		\begin{tabular}{cccc|ccc|ccc}
			\hline 
			\multicolumn{4}{c|}{Loss functions}  & \multicolumn{3}{c|}{$\mu_{test} = 0.4$} & \multicolumn{3}{c}{ $\mu_{test}$ from reference}\\
			\hline 
			$L_{1}$+SSIM &$L_{color}$  &$L_{brightness}$  &$L_{structure}$ &PSNR$\uparrow$  &SSIM$\uparrow$ &CIEDE2000$\downarrow$ &PSNR$\uparrow$  &SSIM$\uparrow$ &CIEDE2000$\downarrow$\\
			\hline 
			\checkmark &   &  &  &22.62  &0.84 &8.67  &26.26  &0.87 &6.20 \\  
			&\checkmark  &\checkmark  &\checkmark  &20.54  &0.83 &10.60  &21.22  &0.84 &9.85  \\ 
			\checkmark &\checkmark  &  &  &22.88  &0.85 & 8.37 &26.49  &0.87 &5.92    \\
			\checkmark &   &\checkmark  &  &\textbf{23.72}  &0.85 &\textbf{7.90}  &\textbf{26.80}  &0.87 &5.86  \\  
			\checkmark &   &  &\checkmark  &23.28  &\textbf{0.86} &8.14 &26.75  &\textbf{0.88} & \textbf{5.84}  \\ 
			\checkmark &\checkmark  &\checkmark  &\checkmark  &23.50  &\textbf{0.86} &8.28  &26.61  &\textbf{0.88} &6.06  \\ 
			\hline 
		\end{tabular}
	}
	\label{table_psnr_ssim}
\end{table*}

Furthermore, as shown in Table \ref{table_psnr_ssim}, all three loss functions based on relative information can lead to improvements in color restoration performance. Among them, the proposed $L_{brightness}$ and $L_{structure}$ demonstrate better color restoration performance than the commonly used $L_{color}$. However, when we use all three loss functions simultaneously, the performance of the CIEDE2000 metric does not reach its optimum. This may be due to conflicts between the assumptions of the models used by the three loss functions. The same issue is also reflected in the PSNR metric. When we only add $L_{brightness}$ to the loss functions, the PSNR value can reach $23.72$ and $26.80$, while when we add all three loss functions simultaneously, the PSNR value can only reach $23.50$ and $26.61$.

\par Fig. \ref{figure_comare_5} and \ref{figure_comare_6} show the impact of different $\mu_{test}$ values on the enhancement results. It can be seen that, as $\mu_{test}$ value changes, the enhanced images' brightness change accordingly. This indicates that the strategy of using the parameter $\mu_{test}$ to control the enhancement results is feasible. 
\begin{figure}[htbp]
	\centering
	\includegraphics [width=\linewidth]{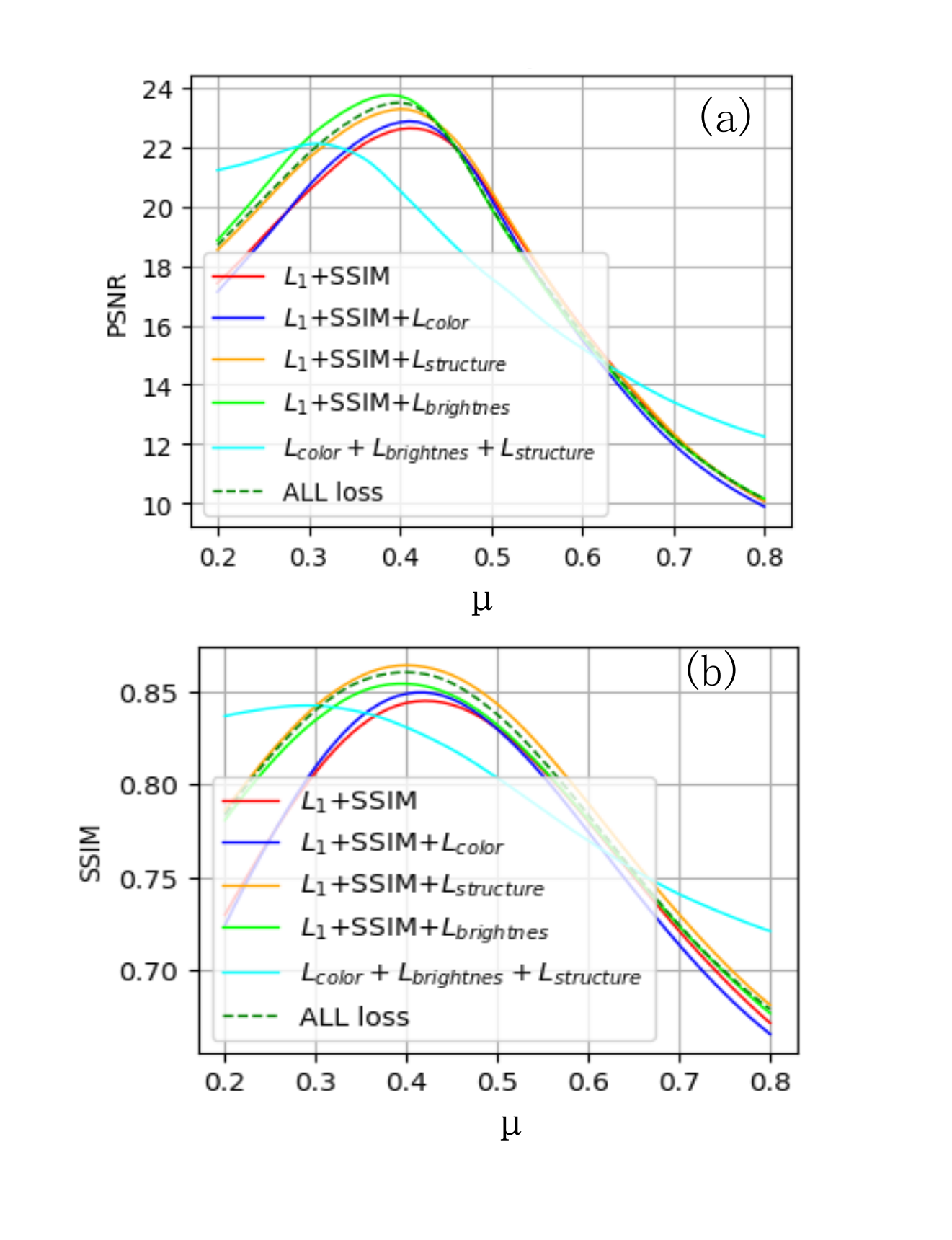}
	\caption{Influence on PSNR and SSIM of different loss functions when changing the $\mu_{test}$ value. (a) The PSNR with different $\mu_{test}$ values. (b) The SSIM with different $\mu_{test}$ value. ALL loss means that all loss functions are added with the same weight (Equation (\ref{eqn12})).}
	\label{fig_psnr_ssim}
\end{figure}
The brightness differences have a significant influence on PSNR and SSIM values. Fig. \ref{fig_psnr_ssim} shows the changes in SSIM and PSNR values with different $\mu_{test}$ values and loss functions. It can be seen that, when the network is trained without $L_{1}$ and SSIM loss, it shows more stable performance in terms of both SSIM and PSNR. If the training losses include $L_{1}$ and SSIM loss, the highest values of both SSIM and PSNR are achieved when $\mu_{test}$ is close to 0.4. This is due to the fact that the artificially selected reference images in the LOL-V2 dataset have a mean value of 0.41 in their V channels. 
There are very few images' V channels with a mean brightness above 0.5, which causes SSIM and PSNR drop significantly when the value of $\mu_{test}$ is higher. 

\par Moreover, in Fig. \ref{fig_psnr_ssim}(b), we can see that the network trained with $L_{1}$, SSIM and $L_{structure}$ loss achieves the highest SSIM. In Fig. \ref{fig_psnr_ssim}(a), the network trained with $L_{1}$, SSIM and $L_{brightness}$ achieved the highest PSNR. This demonstrates the effectiveness of $L_{structure}$ and $L_{brightness}$ in improving structure and brightness restoration, respectively.

\begin{table*}
	\centering
	\caption{The influence of GFE component and loss functions based on relative information. During training, the input $\mu_{train}$ value can be a constant (e.g., $\mu_{train} = 0.4$) or obtained through the reference(e.g., $\mu_{train}$ equals the mean value of the reference's V channel in this table). During testing, the input $\mu_{test}$ value is a constant for all images in this table ($\mu_{test} = 0.4$). Relative losses represents $L_{color} + L_{brightness} + L_{structure}$ }
	\label{table1}
	\resizebox{0.8\textwidth}{!}{
		\begin{tabular}{cc|c|cc|cc}
			\hline 
			\multicolumn{2}{c|}{ Loss functions}  & \multicolumn{1}{c|}{GFE} & \multicolumn{2}{c|}{$\mu_{test} = 0.4$}  & \multicolumn{2}{c}{LOL-V2}\\
			\cline{1-2}  \cline{4-7} 
			$L_{1}$+SSIM  &Relative losses &component  & $\mu_{train}=0.4$  &$\mu_{train}$ from reference    &PSNR$\uparrow$  &SSIM$\uparrow$\\
			\hline 
			\checkmark    &  &  &  &  &18.32  &0.80 \\  
			\checkmark    &\checkmark  &  &  &  &19.05  &0.82    \\
			\checkmark     &         &\checkmark  &\checkmark &  &20.59  &0.83   \\  
			\checkmark     &         &\checkmark  & &\checkmark  &22.62  &0.84   \\  
			\checkmark     &\checkmark  &\checkmark  & &\checkmark    & \textbf{23.50}  &\textbf{0.86}    \\ 
			\hline 
		\end{tabular}
	}
	\label{table_GFE}
\end{table*}

\noindent\textbf{Contribution of GFE component:} In this ablation study, the network trained with $L_{1}$ and SSIM loss without the GFE component was considered as the baseline model. The effects of adding the GFE component and losses proposed in this paper were then compared and studied. The results are presented in Table \ref{table_GFE}. It should be noted that the input $\mu_{test}$ value is constant for all images during testing in this table.

Table \ref{table_GFE} demonstrates that when we add either the other proposed loss functions or the GFE component to the baseline model, both PSNR and SSIM values show improvement. This provides strong evidence for the effectiveness of the GFE component and the loss functions designed with relative information. The GFE component can capture global brightness information and integrate it into the enhancement process with few parameters, leading to a significant improvement in PSNR by 2.27 dB and SSIM by 0.03 (PSNR: $18.32 \rightarrow 20.59$, SSIM: $0.80 \rightarrow 0.83$), even with fixed $\mu_{test}$ and $\mu_{train}$ values during testing and training.

Furthermore, we observed that when the $\mu_{train}$ value is calculated from the reference images during training, the improvement in PSNR and SSIM values is even more significant (e.g., PSNR: $18.32 \rightarrow 22.63$, SSIM: $0.80 \rightarrow 0.84$). This finding highlights the challenge posed by the one-to-many problem in learning how to remove noise and retain structural information in low-light image enhancement. Therefore, the GFE component, which is a simple and efficient method for connecting the input and output images, can greatly improve the enhancement results by facilitating the learning process.

\begin{table*}
	\centering
	\caption{The influence of combining the $L_{brightness}$ and $L_{structure}$ with RetinexNet on LOL-V1 dataset. In the original paper\citep{wei2018deep}, RetienxNet was trained with LOL-V1 dataset, and the authors used BM3D to denoise the reflectance image.  }
	\label{table1}
	\resizebox{0.65\textwidth}{!}{
		\begin{tabular}{ccc|ccc}
			\hline 
			Original loss  & BM3D  &$L_{brightness} \& L_{structure}$    &PSNR$\uparrow$  &SSIM$\uparrow$ &CIEDE2000$\downarrow$\\
			\hline 
			\checkmark    &             &             &16.79   &0.42 &15.89 \\  
			\checkmark    &\checkmark   &             &17.86   &0.78 &13.76 \\
			\checkmark     &             &\checkmark   &\textbf{18.87}   &\textbf{0.79} &\textbf{12.01} \\  
			\hline 
		\end{tabular}
	}
	\label{table_retinexnet}
\end{table*}

\subsection{Combined With Other Networks}
\begin{figure*}[htbp]
	\centering
	\includegraphics [width=0.9\linewidth]{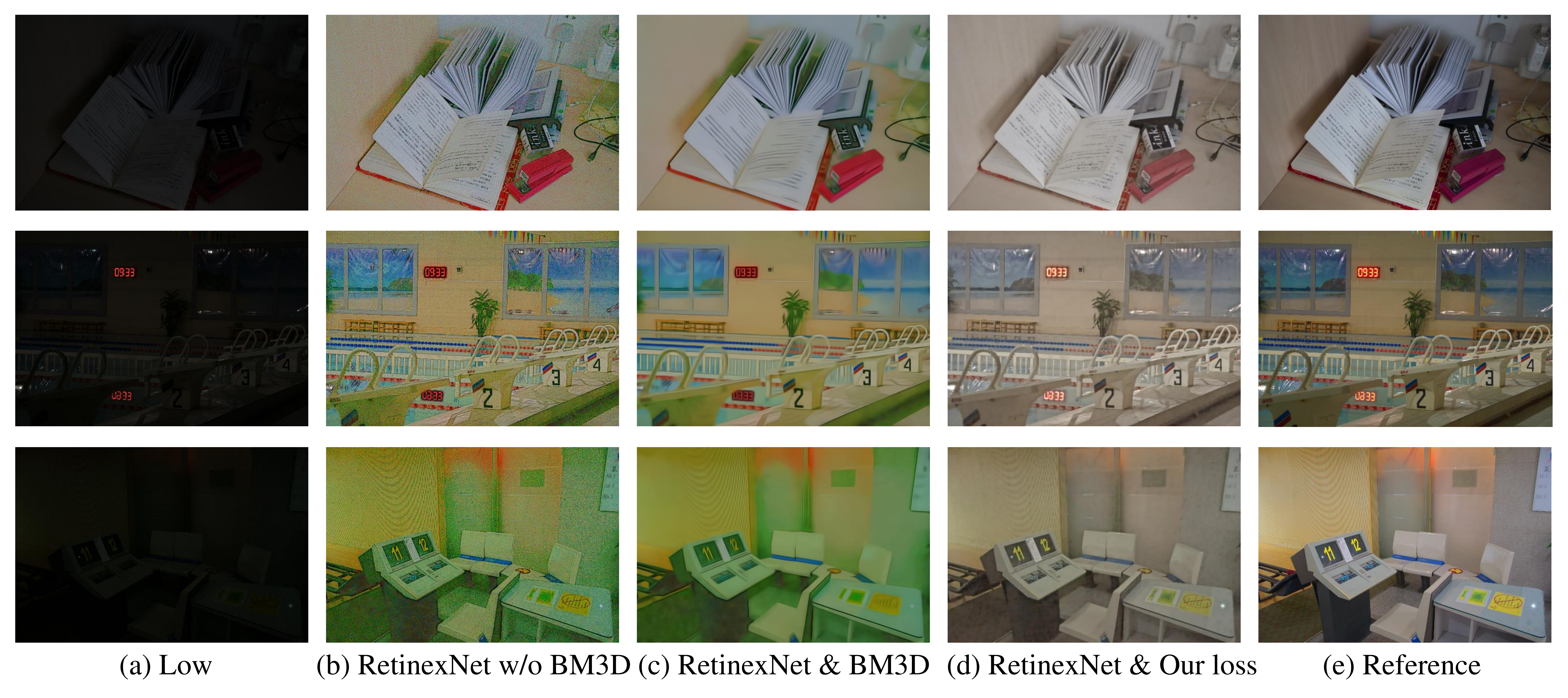}
	\caption{The visual effect of combining the proposed loss functions with RetinexNet. (a) The input low-light images. (b) Results of RetinexNet without BM3D. (c) Results of RetinexNet with BM3D. (d) Results of RetinexNet without BM3D and with $L_{brightness}$ and $L_{structure}$ in the training loss function. (e) Reference images. In the original paper \citep{wei2018deep}, RetinexNet was trained with LOL-V1 dataset and BM3D was used to denoise on the reflectance image. (Best viewed on high-resolution displays with zoom-in.)}
	\label{fig_retinexnet}
\end{figure*}

\begin{figure*}[htbp]
	\centering
	\includegraphics [width=0.9\linewidth]{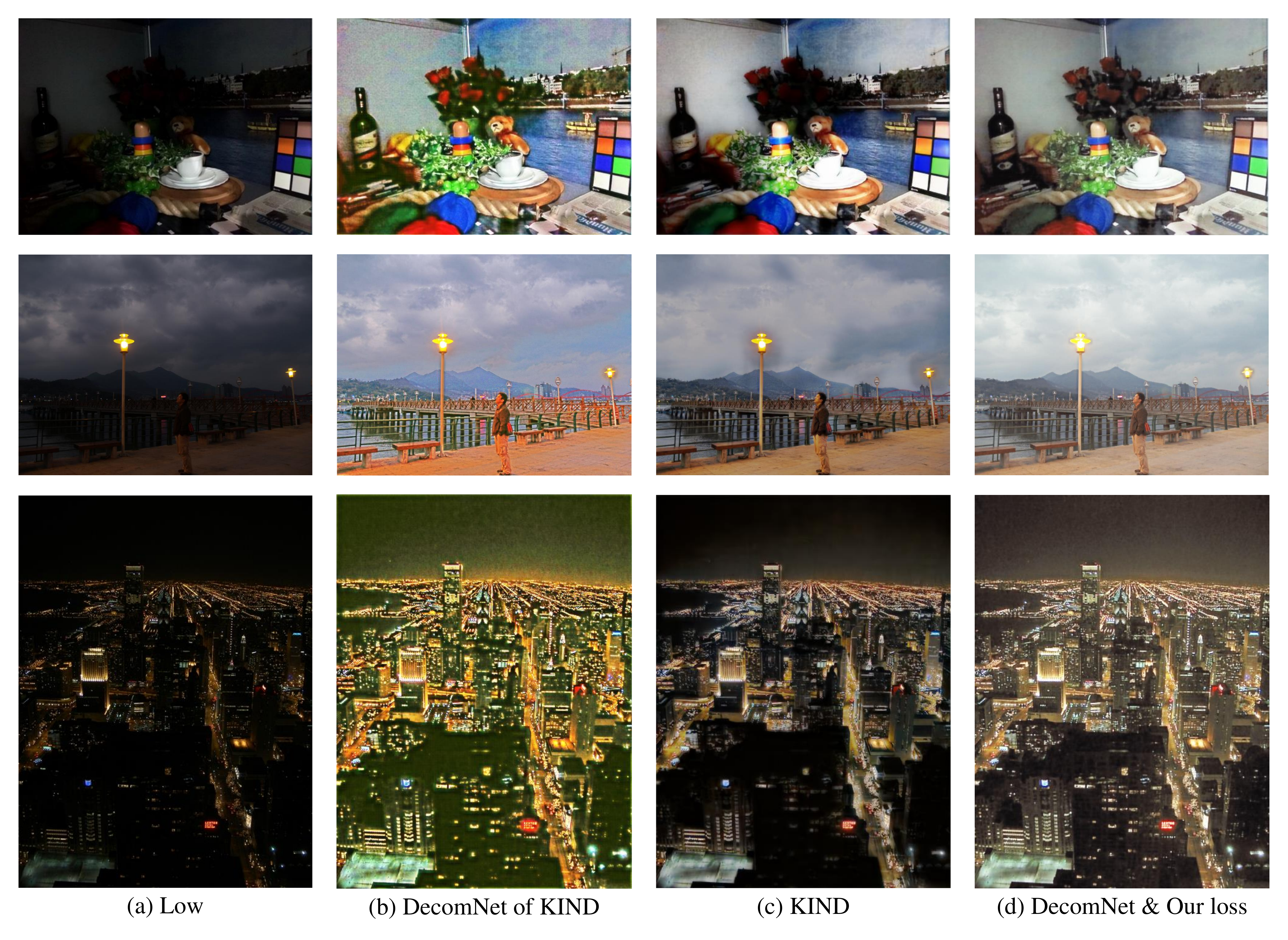}
	\caption{The visual effect of combining the loss function with DecomNet of KIND. KIND has three sub-networks, which are the Decomposition Network, the Restoration Network, and the Illumination Adjustment Network. They are abbreviated as DecomNet, RestorNet, and AdjustNet in this table. (a) The input low-light images. (b) Results of DecomNet in KIND. (c) KIND. (d) DecomNet in KIND with $L_{brightness}$ and $L_{structure}$ in the original loss function. (Best viewed on high-resolution displays with zoom-in.)}
	\label{fig_kind}
\end{figure*}

The proposed $L_{brightness}$ and $L_{structure}$ loss functions can also be combined with other supervised image enhancement methods to improve their performance with fewer operations or parameters . For example, BM3D or additional sub-network is adopted to denoise on the reflectance image in Retinex based low-light image enhancement methods(e.g., RetinexNet \citep{wei2018deep}, KIND \citep{zhang2019kindling}, and KIND++\citep{li2021learning}). However, if we add the proposed two relative loss functions into the training of the network, their performace can be imporved without the additional part. Table \ref{table_retinexnet}, \ref{table_kind} and Fig. \ref{fig_retinexnet}, \ref{fig_kind} show the examples of combining $L_{brightness}$ and $L_{structure}$ with RetinexNet and KIND. 


RetinexNet was trained with LOL-V1 dataset in the original paper \citep{wei2018deep}, and BM3D was used for denoising reflectance images. By adding the two proposed relative loss functions to the training of RetinexNet, the network achieves better PSNR, SSIM, and CIEDE2000 scores than the original RetinexNet and RetinexNet combined with BM3D, as Table \ref{table_retinexnet} shows. Also, as seen in Fig. \ref{fig_retinexnet}, RetinexNet with $L_{brightness}$ and $L_{structure}$ loss functions can significantly reduce noise in the enhanced image (Fig. \ref{fig_retinexnet}(b) and (d)). Compared with the BM3D method (Fig. \ref{fig_retinexnet}(c)), proposed relative loss functions can help preserve more details and restore the color (Fig. \ref{fig_retinexnet}(d)).

KIND consists of three sub-networks: DecomNet, RestorNet, and AdjustNet. By adding the proposed relative loss functions to the training of DecomNet, DecomNet achieved better results than the entire KIND method, as shown in Table \ref{table_kind}. At that time, the number of parameters can be reduced by nearly $95\%$. 
\begin{table*}
	\centering
	\caption{The influence of combining the $L_{brightness}$ and $L_{structure}$ with KIND on LOL-V1 dataset. In the original paper\citep{zhang2019kindling}, KIND was trained with LOL-V1 dataset. KIND has three sub-networks, which are the Decomposition Network, the Restoration Network, and the Illumination Adjustment Network. They are abbreviated as DecomNet, RestorNet, and AdjustNet in this table.}
	
	\resizebox{0.9\textwidth}{!}{
		\begin{tabular}{ccc|ccccc}
			\hline 
			DecomNet  & RestorNet$\&$AdjustNet  &$L_{brightness} \& L_{structure}$    &PSNR$\uparrow$  &SSIM$\uparrow$ &CIEDE2000$\downarrow$ & Params(M)$\downarrow$  \\
			\hline 
			\checkmark    &             &             &15.68   &0.50 &18.73 &0.43  \\  
			\checkmark    &\checkmark   &             &17.65   &0.78 &12.49 &8.21  \\
			\checkmark     &             &\checkmark   &\textbf{18.94}   &\textbf{0.80} &\textbf{11.93} &0.43 \\  
			\hline 
		\end{tabular}
	}
	\label{table_kind}
\end{table*}
Figure \ref{fig_kind} show some images of combining the relative loss functions with DecomNet of KIND. It can be seen that, the relative loss functions have positive impact on noise reduction, color correction, and detail preservation. Thus, we can obtain comparable or superior results using a simpler network structure with the two proposed relative loss functions.

\section{Conclusion}
In this paper, we have demonstrated that by using efficient global feature extraction and the proposed relative loss functions, a simple network structure can be employed to achieve image enhancement that is comparable to, or even better than, the current state-of-the-art (SOTA) methods with much less running time. Moreover, the Global Feature Extraction component and loss functions can be combined with other low-light image enhancement techniques to enhance objective evaluation indicators such as PSNR and SSIM. The experimental results show the effectiveness and advantages of our method for low-light image enhancement. However, our approach still has some limitations, such as the dependence of the final enhancement result on the desired brightness parameter $\mu_{test}$ and the requirement for paired data during training. To address the first issue, parameters can be automatically selected, as demonstrated in our previous work \citep{fu2020learning}, or the contrast of the enhanced image can be further adjusted through GAMMA Correction or other local tone mapping techniques \citep{zeng2020learning}. In future research, we aim to improve the robustness of the proposed method and explore unsupervised approaches for network training.

\bmhead{Acknowledgments}
The authors would like to thank for National Natural Science Foundation of China (No. 62027283 and 61775048), Aeronautical Science Foundation of  China(No.2022Z071077002) and Shenzhen Fundamental Research Program (No. JCYJ2020109150808037) Funded.

\section*{Declarations}

\begin{itemize}
\item The authors have no competing interests to declare that are relevant to the content of this article.
\item The code is available at \url{https://github.com/hitzhangyu/FLW-Net}
\item The data used in the paper is sourced from publicly published papers or publicly available websites. However, if there are reasonable justifications, it is also possible to request the data from the corresponding authors.
\item Y.Zhang designed the workflow, and wrote the manuscript. J.Wu., Y.Wang., Y.Li., and R.Fu. provided some ideas and solved some problems in the work, and contributed to the manuscript. Y.Xu. and Y.Guo provide some ideas and contributed to the manuscript. X.Xiao, and C.Wang. supervised the work and contributed to the manuscript.
\end{itemize}


\bibliography{sn-bibliography}


\end{document}